\documentclass{article}
\usepackage[preprint]{neurips_2025}

\usepackage{natbib} 
\usepackage[colorlinks=true, citecolor=blue, linkcolor=blue, urlcolor=blue, anchorcolor=blue]{hyperref}
\usepackage{graphicx}
\usepackage{subcaption}
\usepackage[utf8]{inputenc} 
\usepackage[T1]{fontenc}    
\usepackage{breakurl}     
\usepackage{listings}
\usepackage{caption}
\usepackage{booktabs}       
\usepackage{threeparttable}
\usepackage{listings}
\usepackage{placeins}
\usepackage{amsfonts}       
\usepackage{amsmath}
\usepackage{amsthm}
\usepackage{xurl}
\usepackage{amssymb}
\usepackage{nicefrac}       
\usepackage{microtype}      
\usepackage[font=small,labelfont=bf]{caption}
\usepackage{enumitem}
\usepackage{wrapfig}
\usepackage{listings}
\usepackage{caption}
\usepackage{multirow}
\usepackage{multirow}
\usepackage[most]{tcolorbox}
\usepackage[normalem]{ulem}
\usepackage{xspace}
\usepackage{float}
\usepackage{tabularx}
\usepackage{booktabs}
\usepackage[normalem]{ulem}
\usepackage{soul}   
\usepackage{svg}
\usepackage{array}
\usepackage{longtable}
\usepackage{adjustbox}
\usepackage{makecell}
\usepackage{pifont}
\usepackage{pdflscape}

\usepackage{xcolor}

\useunder{\uline}{\ul}{}

\definecolor{mygreybg}{gray}{0.95}

\sethlcolor{mygreybg}

\title{Voxtral TTS}

\begin{document}
\maketitle
\vspace{-0.1in}
\begin{center}
\vspace{-45pt}
\centering
\includegraphics[width=0.8\linewidth,keepaspectratio]{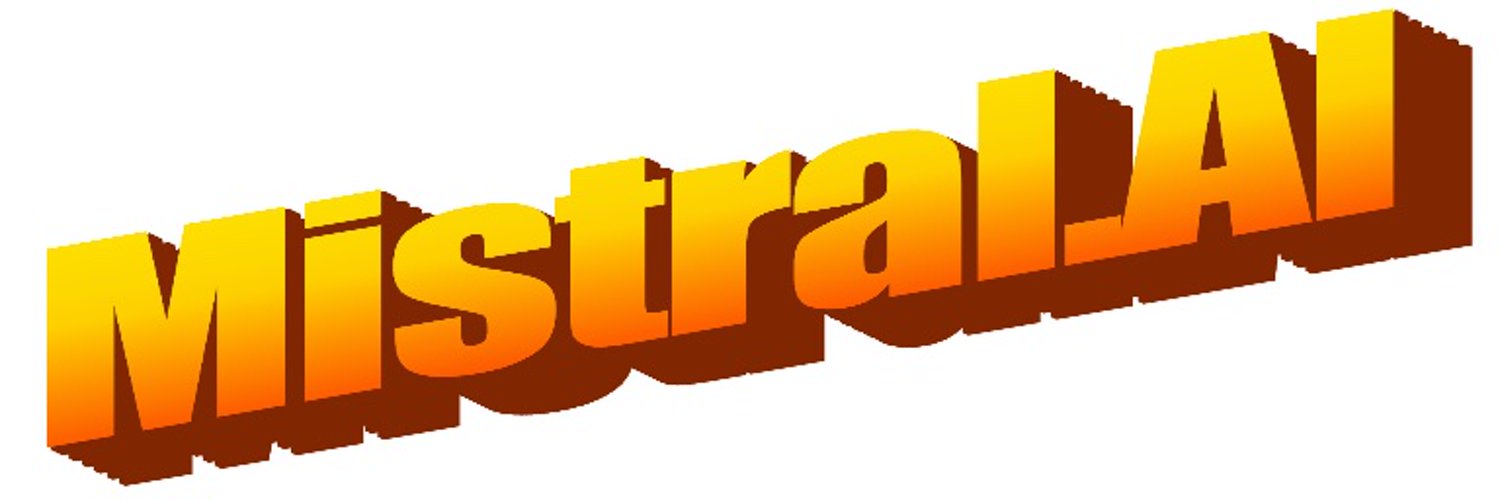}
\end{center}

\begin{abstract}
We introduce Voxtral TTS, an expressive multilingual text-to-speech model that generates natural speech from as little as 3 seconds of reference audio. Voxtral TTS adopts a hybrid architecture that combines auto-regressive generation of semantic speech tokens with flow-matching for acoustic tokens. These tokens are encoded and decoded with Voxtral Codec, a speech tokenizer trained from scratch with a hybrid VQ-FSQ quantization scheme. In human evaluations conducted by native speakers, Voxtral TTS is preferred for multilingual voice cloning due to its naturalness and expressivity, achieving a 68.4\% win rate over ElevenLabs Flash v2.5. We release the model weights under a CC BY-NC license.
\end{abstract}
\begin{center}

\begin{tabular}{@{} l l @{}}
\small{\textbf{Webpage:}}  & \scriptsize{\url{https://mistral.ai/news/voxtral-tts}} \\
{\small{\textbf{Model weights:}}}  & \scriptsize{\url{https://huggingface.co/mistralai/Voxtral-4B-TTS-2603}} \\
\end{tabular}
\end{center}

\begin{figure*}[h]
\begin{center}
\includegraphics[width=\linewidth, trim={0 35 0 0},clip]{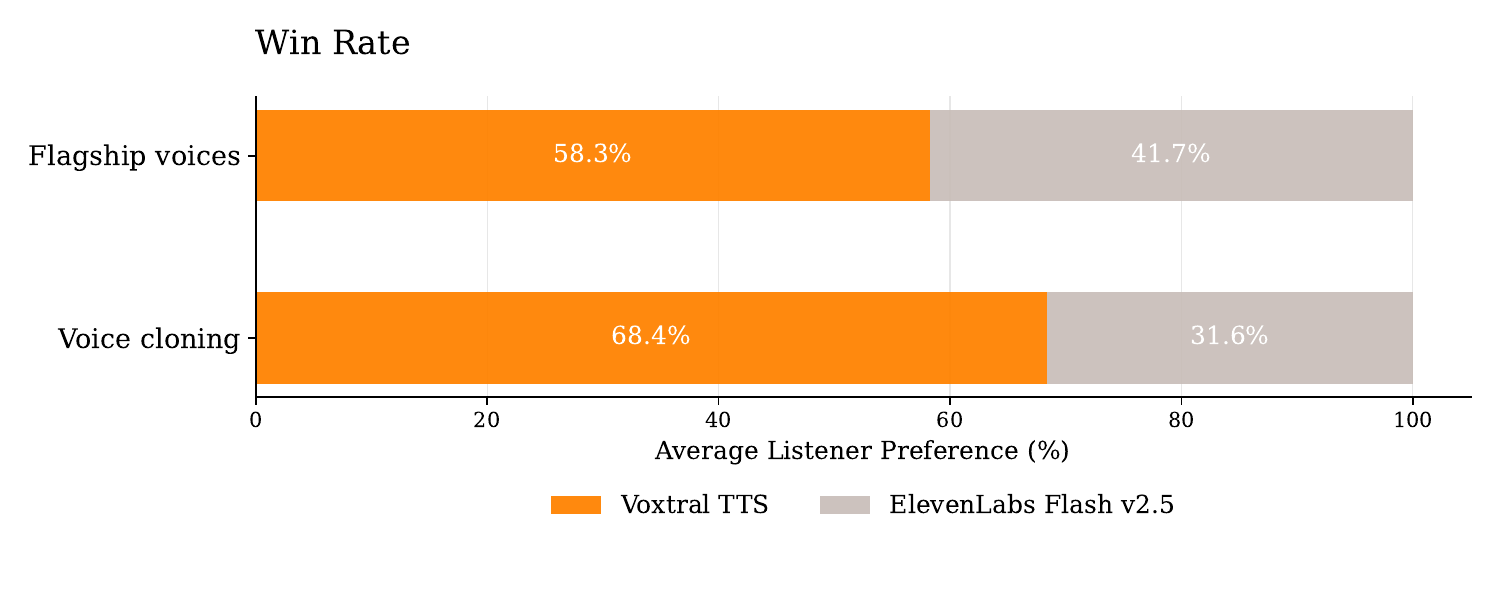}
\end{center}
\caption{\textbf{Voxtral TTS is preferred to ElevenLabs Flash v2.5 in human evaluations.} We plot the win rate for Voxtral TTS against ElevenLabs Flash v2.5 in human evaluations across two categories. For flagship voices, we use the default voices for each model and 77 unique text examples. In the voice cloning set-up, we provide a short audio reference clip and 60 text prompts. In both categories, human annotators blindly rate which audio is better between the two models. Voxtral TTS is preferred in 58.3 and 68.4\% of instances.}
\label{fig:win-rate}
\end{figure*}

\section{Introduction}
Natural and expressive text-to-speech (TTS) remains a cornerstone of flexible human-computer interactions, with applications spanning virtual assistants, audiobooks, and accessibility tools. While recent neural TTS models achieve strong intelligibility, capturing the nuances and expressivity of human speech remains an open challenge, particularly in the zero-shot voice setting. 

Recent zero-shot TTS systems typically condition generation on discrete speech tokens extracted from a short voice prompt, enabling generalization to unseen speakers and natural synthesis across long sequences~\citep{audiolm, vall_e}. In parallel, diffusion and flow-based models are effective for modeling rich acoustic variation in speech generation~\citep{gradtts, voicebox}. Recent speech codecs demonstrate that speech can be factorized into a low-rate semantic stream and a higher-rate acoustic stream~\citep{defossez2024moshi}. Hierarchical generators such as Moshi already exploit this structure using a temporal transformer over timesteps and a depth transformer over codec levels. However, acoustic generation in these systems remains depth-wise autoregressive. For TTS, this raises the question whether the dense acoustic component must be modeled auto-regressively at all, or whether it can instead be generated more effectively with a conditional continuous model.


In this work, we introduce Voxtral TTS, a multilingual zero-shot TTS system built around a representation-aware hybrid architecture. A voice prompt is tokenized through Voxtral Codec, a low-bitrate speech tokenizer with an ASR-distilled semantic token and finite scalar quantized (FSQ) acoustic tokens~\citep{fsq}. Given this factorized representation, a decoder-only transformer auto-regressively predicts the semantic token sequence, while a lightweight flow-matching model predicts the acoustic tokens conditioned on the decoder states. This design combines the strengths of auto-regressive modeling for long-range consistency with continuous flow-matching for rich acoustic detail. We adapt Direct Preference Optimization (DPO)~\citep{dpo} to this hybrid discrete-continuous setting by combining a standard preference objective over semantic token generation with a flow-based preference objective for acoustic prediction~\citep{flow-dpo}.

Voxtral TTS supports 9 languages, supports voice prompts as short as 3 seconds, and is designed for low-latency streaming inference. Across automatic evaluations on SEED-TTS~\citep{seedtts} and MiniMax-TTS~\citep{minimaxtts}, it achieves strong intelligibility and naturalness, beating ElevenLabs v3 on speaker similarity scores. In human evaluation for multilingual zero-shot voice cloning, it is preferred over ElevenLabs Flash v2.5 with a 68.4\% win rate, while remaining competitive with strong proprietary systems on expressive flagship-voice evaluations.


\vspace{-0.025in}
\section{Modeling}
\label{sec:modeling}
\begin{figure*}
\begin{center}
\includegraphics[width=0.9\linewidth,keepaspectratio, trim={0 0 0 2.5}, clip]{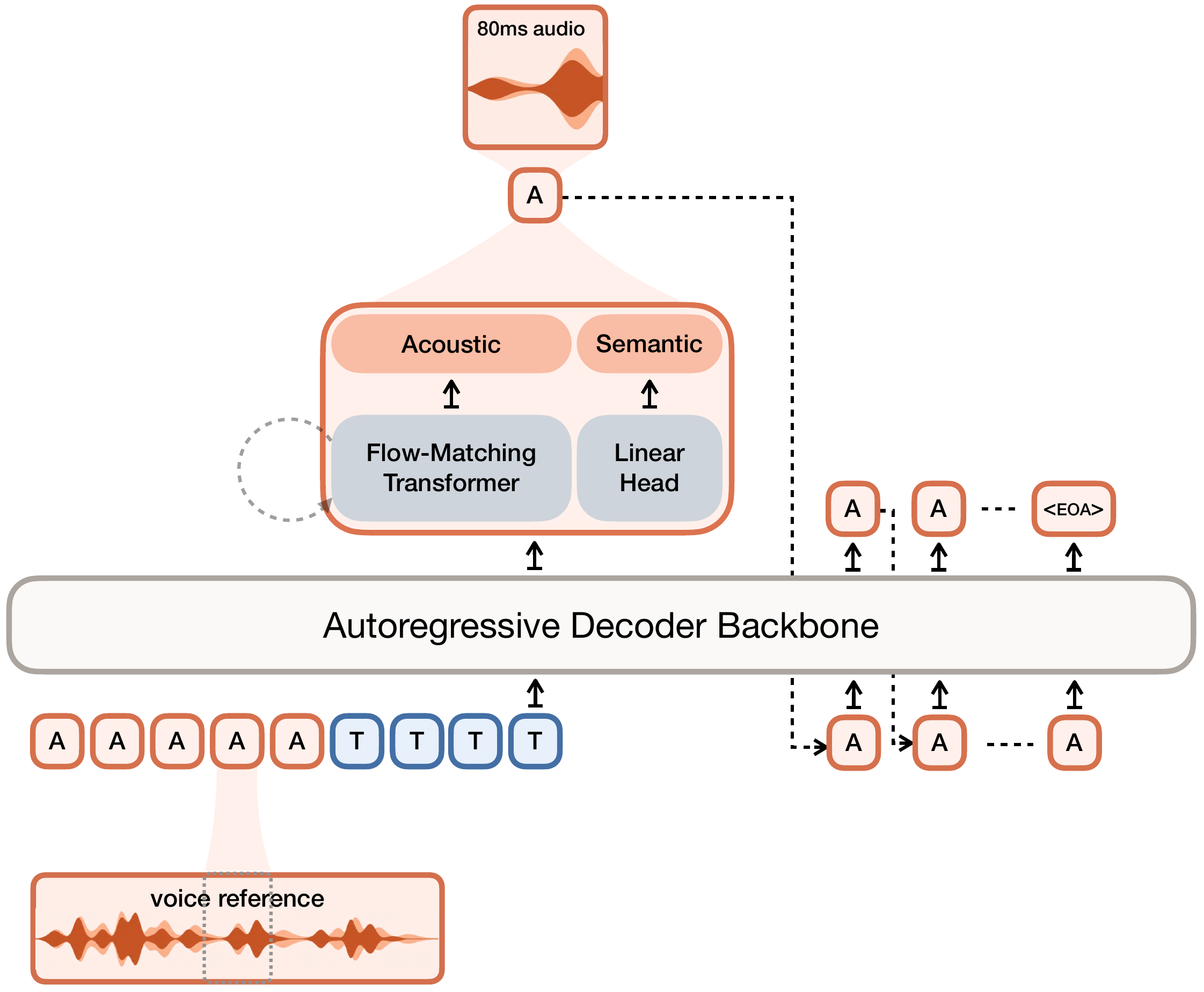}
\end{center}
\caption{\textbf{Architecture overview of Voxtral TTS.} A voice reference ranging from 3s-30s is fed to the Voxtral Codec encoder to obtain audio tokens at a frame rate of 12.5 Hz. Each audio frame (labeled \textbf{A}) consists of a semantic token and acoustic tokens. The voice reference audio tokens along with the text prompt tokens (labeled \textbf{T}) are fed to the decoder backbone. The decoder auto-regressively generates a sequence of semantic tokens until it reaches a special End of Audio token (\textbf{<EOA>}). At each timestep, the semantic token from the decoder backbone is fed to a flow-matching transformer, which is run multiple times to predict the acoustic tokens. The semantic and acoustic tokens are fed to the Voxtral Codec decoder to obtain the generated waveform.}
\label{fig:voxtral-tts}
\end{figure*}

Figure~\ref{fig:voxtral-tts} highlights the architecture of Voxtral TTS. It consists of a novel audio codec---Voxtral Codec---which encodes a reference voice sample into audio tokens consisting of semantic and acoustic tokens. The audio tokens are combined with text tokens to form the input to the LM decoder backbone. To generate speech, the decoder backbone auto-regressively generates semantic token outputs. A flow-matching transformer generates the acoustic tokens. The codec decoder maps the output tokens to the corresponding audio waveform.

\vspace{-0.025in}
\subsection{Voxtral Codec}

\begin{figure*}
\begin{center}
\includegraphics[width=\linewidth,keepaspectratio]{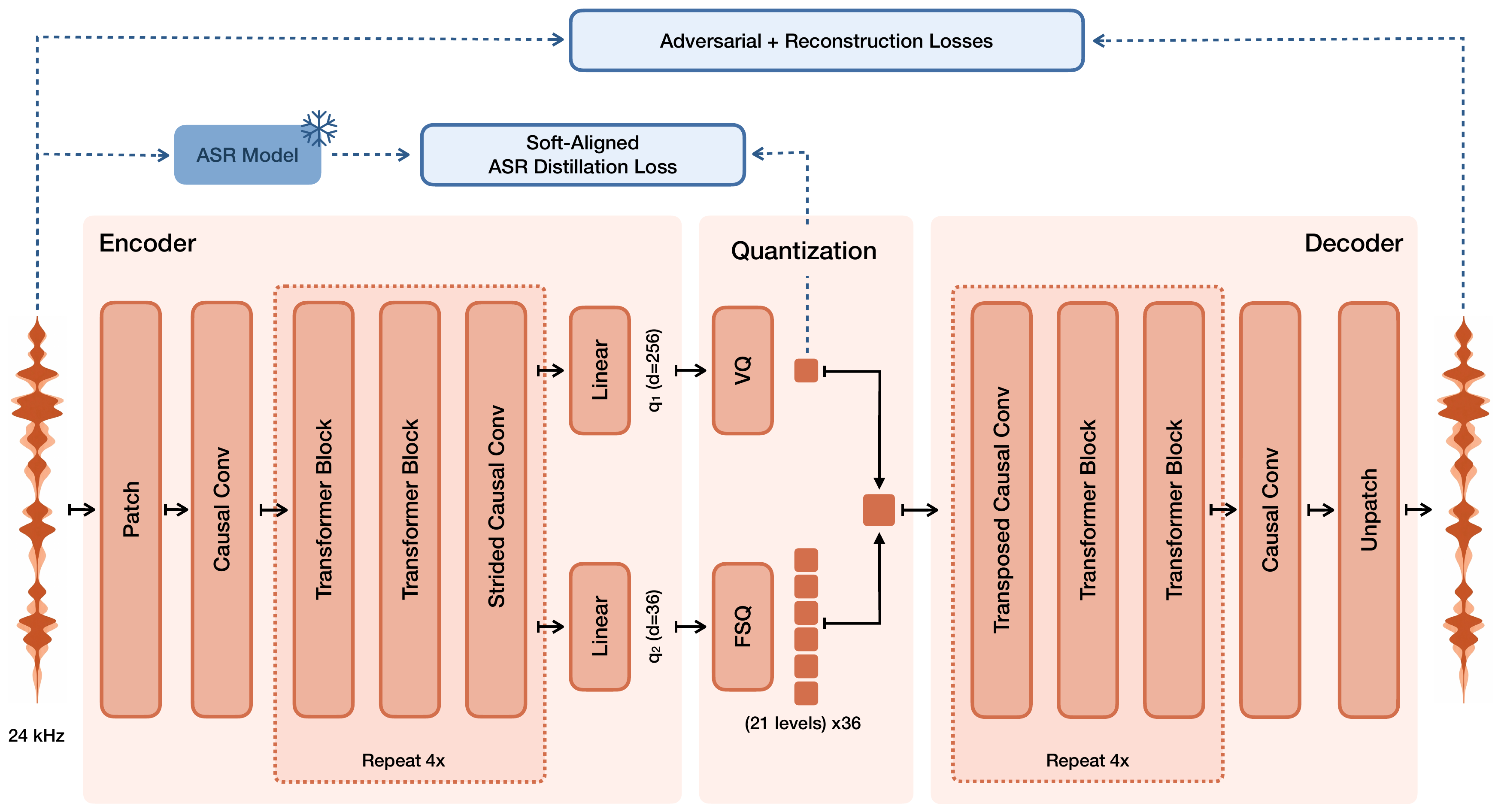}
\end{center}
\caption{\textbf{Architecture overview and training of Voxtral Codec.} It consists of a split semantic VQ codebook and acoustic FSQ codebooks. Both semantic and acoustic tokens are combined for reconstruction. The semantic token has an additional distillation loss from a supervised ASR model.}
\label{fig:voxtral-codec}
\end{figure*}

\looseness=-1 Voxtral Codec is a convolutional--transformer autoencoder~\citep{defossez2022high} that compresses raw 24~kHz mono waveforms into 12.5~Hz frames of 37 discrete tokens (1 semantic + 36 acoustic), achieving a total bitrate of 2.14~kbps. These tokens serve as the input audio representation to Voxtral TTS. Through a novel combination of architectural and training objective improvements, Voxtral Codec outperforms existing baselines such as Mimi~\citep{defossez2024moshi}, with results presented in Section~\ref{sec:codec-results}.

\paragraph{Waveform Autoencoder.} Inspired by prior works on transformer-based audio codecs~\citep{parker2024scaling,wu2024ts3}, our audio tokenizer operates on ``patchified'' waveforms. A 24\,kHz mono input waveform is chunked into non-overlapping patches of 240 samples, yielding a 100~Hz input to the encoder. The 100~Hz input frames are first projected to 1024-dimensional embeddings via a causal convolution with kernel size 7. The embeddings are then forwarded through 4 encoder blocks, each comprising:
\begin{itemize}
    \item A 2-layer causal self-attention transformer with sliding window attention (window sizes $16 \rightarrow 8 \rightarrow 4 \rightarrow 2$, halved at each downsampling stage), ALiBi positional bias~\citep{press2021train}, QK-norm, and LayerScale~\citep{touvron2021going} initialized at 0.01.
    \item A causal CNN layer. In the first three blocks, the CNN downsamples by 2$\times$ (stride 2), yielding a cumulative 8$\times$ reduction from 100~Hz to 12.5~Hz. In the fourth block, the CNN has stride 1 and projects the 1024-dimensional representation to a 292-dimensional latent space.
\end{itemize}

The 292-dimensional latent is subsequently quantized to audio tokens (detailed below).
The decoder mirrors the encoder in reverse: a causal CNN first projects the 292-dimensional latent back to 1024 dimensions, followed by 4 blocks each containing a transposed CNN (for 2$\times$ upsampling) and a 2-layer causal self-attention transformer, gradually restoring the 12.5~Hz latent to 100~Hz. A final causal convolution with kernel size 7 maps from 1024 dimensions back to the patch size of 240 samples to reconstruct the waveform.

\vspace{-0.02in}

\paragraph{Representation Quantization.}

The 292-dimensional latent is split into a 256-dimensional \emph{semantic} component and a 36-dimensional \emph{acoustic} component, which are quantized independently:
\begin{itemize}
    \item The semantic component is quantized through a learned vector quantizer (VQ;~\citep{van2017neural}) with a codebook of size 8192. During training, VQ is applied with 50\% probability; the remaining samples pass through unquantized. 
    \item Each of the 36 acoustic dimensions is passed through a $\tanh$ activation and independently quantized to 21 uniform levels via finite scalar quantization (FSQ;~\citep{fsq}). During training, we apply dither-style FSQ~\citep{parker2024scaling}: 50\% of samples are quantized with FSQ, 25\% receive uniform noise of magnitude $1/L$ (where $L{=}21$ is the number of levels), and 25\% pass through unquantized.
\end{itemize}
The total bitrate is $12.5 \times (\log_2 8192 + 36 \times \log_2 21) \approx 2.14$~kbps.

\looseness=-1 \paragraph{Semantic Token Learning.} To better incorporate the semantic content of speech into the semantic tokens, we adopt an auxiliary ASR distillation loss. Unlike prior works that learn ``semantic'' tokens by distilling self-supervised speech representations~\citep{zhang2023speechtokenizer,defossez2024moshi}, which are more \textit{phonetic} than semantic~\citep{liu2024revisiting}, we distill from a supervised ASR model. This has been shown to produce more effective semantic representations~\citep{vashishth2024stab}.

A frozen Whisper~\citep{radford2023robust} model is run auto-regressively on the input audio to generate decoder hidden states and cross-attention weights. The post-VQ semantic embeddings are linearly projected to match the Whisper hidden dimension and then aligned to the decoder hidden states from the last decoder layer using a cosine distance loss:
\begin{equation}
    \mathcal{L}_\text{ASR} = 1 - \frac{1}{L}\sum_{l=1}^{L} \frac{\tilde{\boldsymbol{z}}_l \cdot \boldsymbol{h}_l}{\lVert \tilde{\boldsymbol{z}}_l \rVert \, \lVert \boldsymbol{h}_l \rVert}, \qquad \tilde{\boldsymbol{z}}_l = \sum_{f=1}^{F} A_{l,f} \, \boldsymbol{z}_f
\end{equation}
where $\boldsymbol{z}_f$ are the projected post-VQ semantic embeddings at codec frame $f$, $\boldsymbol{h}_l$ are the last-layer decoder hidden states from Whisper at token position $l$, and $A \in \mathbb{R}^{L \times F}$ is a soft alignment matrix derived from a subset of Whisper's cross-attention heads identified as best correlating with word-level timestamps via dynamic time warping (DTW)~\citep{berndt1994dtw}. To compute $A$, the cross-attention weights from these heads are normalized across the decoder token dimension, median-filtered, and averaged over heads. The resulting matrix is linearly interpolated along the encoder frame axis to match the codec frame rate (12.5~Hz), so that $\tilde{\boldsymbol{z}}_l$ is the attention-weighted sum of codec embeddings aligned to the $l$-th decoder token.

This design allows the tokenizer to learn text-aligned semantic tokens without requiring an external forced aligner or paired transcripts, since the alignment is derived implicitly from Whisper's cross-attention weights. Distilling from continuous hidden states rather than hard transcript labels provides richer supervision, including model confidence and phonetic similarities.

\paragraph{Adversarial Training.}

A multi-resolution discriminator with 8 STFT sizes (2296, 1418, 876, 542, 334, 206, 126, 76) is trained along with the codec. Each discriminator is trained as a binary classifier between real audios $\boldsymbol{x}$ and reconstructed audios $\boldsymbol{\hat{x}}$ using a hinge loss. An $L_1$-based feature-matching loss is computed on the activations of every layer of each discriminator:
\vspace{-0.025in}

\begin{equation}
    \mathcal{L_\text{feature}}(\boldsymbol{x}, \boldsymbol{\hat{x}}) = \frac{1}{MN}\sum_{m=1}^M\sum_{n=1}^N \lVert D_n^m(\boldsymbol{x}) - D_n^m(\boldsymbol{\hat{x}})\rVert_1
\end{equation}

Here, $D_n^m$ denotes the $m$-th layer of the $n$-th discriminator, where each of the $N$ discriminators has $M$ layers. Following~\cite{defossez2024moshi,parker2024scaling}, we use this feature-matching loss \emph{in place of} the standard GAN generator loss, as the evolving discriminator features provide an increasingly discriminative reconstruction signal throughout training.

\paragraph{Training Objective.} Voxtral Codec is trained end-to-end with the following losses:
\begin{equation}
    \alpha \mathcal{L_\text{feature}} + \beta \mathcal{L_\text{ASR}} + \gamma \mathcal{L_\text{L1}} + \gamma \mathcal{L_\text{STFT}} + \delta \mathcal{L_\text{commit}}
\end{equation}
\looseness=-1 where $\alpha{=}1.0$, $\beta{=}1.0$, $\gamma{=}0.9999^t$ (with $t$ the current training step), and $\delta{=}0.1$. $\mathcal{L}_\text{L1}$ is the $L_1$ distance between the original and reconstructed waveforms, and $\mathcal{L}_\text{STFT}$ is an $L_1$ loss on their STFT magnitudes. Both reconstruction losses share the same exponential decay schedule $\gamma$, which bootstraps learning early in training and diminishes their influence as the adversarial signal strengthens~\citep{parker2024scaling}. $\mathcal{L}_\text{commit} = \lVert \boldsymbol{z}_e - \mathrm{sg}(\boldsymbol{z}_q) \rVert_2^2$ is the VQ commitment loss~\citep{van2017neural}, where $\mathrm{sg}$ denotes the stop-gradient operator.

\begin{table}
\centering
\begin{threeparttable}
\caption{Key hyperparameters of the Voxtral Codec.}
\small
\label{tab:codec_arch}
\begin{tabular}{lll}
\toprule
\textbf{} & \textbf{Parameter} & \textbf{Value} \\
\midrule

\multicolumn{3}{l}{\textit{Input / Preprocessing}} \\
& Sampling rate & 24000 \\
& Patch size & 240 \\
\midrule

\multicolumn{3}{l}{\textit{AutoEncoder}} \\
& Encoder patch projection kernel size & 7 \\
& Encoder patch projection dimension & 1024 \\
& Encoder transformer layers\tnote{1} & $2\rightarrow2\rightarrow2\rightarrow2$ \\
& Encoder sliding window size & $16\rightarrow8\rightarrow4\rightarrow2$ \\
& Encoder conv kernels & $4\rightarrow4\rightarrow4\rightarrow3$ \\
& Encoder conv strides & $2\rightarrow2\rightarrow2\rightarrow1$ \\
\multicolumn{3}{c}{\textit{(Decoder flips all $\rightarrow$ to $\leftarrow$ and uses transposed convolutions)}}\\
\midrule

\multicolumn{3}{l}{\textit{Discrete bottleneck}} \\
& Semantic VQ\tnote{2}~~ codebook size & 8192 \\
& Acoustic FSQ\tnote{3}~~codebook count$\times$size & $36\times21$\\
\midrule

\multicolumn{3}{l}{\textit{Discriminator}} \\
& FFT sizes & 2296, 1418, 876, 542, 334, 206, 126, 76 \\
& Channels & 256 \\
\bottomrule
\end{tabular}

\begin{tablenotes}
\footnotesize
\item[1] {\scriptsize For training stability, we use LayerScale with initial scale of 0.01 and QK normalization with $\epsilon = 10^{-6}$.}
\item[2] {\scriptsize During training, VQ is applied with 50\% probability.}
\item[3] {\scriptsize During training: 50\% quantized with FSQ, 25\% dithered (uniform noise of magnitude $1/L$), 25\% unquantized.}
\end{tablenotes}

\end{threeparttable}
\end{table}
Table~\ref{tab:codec_arch} presents a summary of the Voxtral Codec configuration.
The full model has approximately 300M parameters. All decisions are ablated and the final configuration achieves stable optimization with the best audio quality.

\subsection{Decoder Backbone}

The decoder backbone of Voxtral TTS follows the architecture of Ministral 3B~\citep{liu2026ministral}, an auto-regressive decoder-only transformer. 
The input sequence consists of voice reference audio tokens followed by text tokens, from which the output audio tokens are auto-regressively generated. Each audio frame is represented by 37 discrete tokens (1 semantic, 36 acoustic). Each codebook has its own embedding lookup table (8192 entries for semantic and 21 for each acoustic), which are summed to produce a single embedding per audio frame.

The decoder backbone generates a sequence of hidden states. A linear head projects each hidden state $h$ to logits over the semantic codebook vocabulary (8192 entries plus a special End of Audio (\textbf{<EOA>}) token), trained with a standard cross-entropy loss. To predict the acoustic tokens, $h$ is fed to a flow-matching transformer, described in Section~\ref{sec:fmtfm}. The float-valued outputs of the flow-matching transformer are discretized before the next AR step to maintain a fully discrete token interface.

\subsection{Flow-Matching Transformer}\label{sec:fmtfm}
To predict the acoustic tokens, a flow-matching (FM) transformer operates independently on the hidden state $h$ from each generation step in the decoder backbone. We model acoustic tokens in continuous space to leverage the smooth velocity field of FM, and discretize only at the output to interface with the AR backbone's discrete token vocabulary.

The FM transformer consists of a bidirectional 3-layer transformer with the same width as the decoder backbone. It models the velocity field that transports Gaussian noise ($x_0$) to acoustic embedding ($x_1$) over a series of function evaluation steps $0 \leq t \leq 1$. It receives as input $h$, the current function evaluation step $t$ encoded as a sinusoidal embedding, and the current acoustic embedding $x_t \in \mathbb{R}^{36}$. We use a separate projection layer for each input $h$, $t$ and $x_t$, because the scale of activations are different for each one. We also ablated providing conditioning using DiT style adaptive LayerNorm (AdaLN) layers~\citep{dit}, but found our approach superior. 


During training, the hidden state is dropped out 10\% of the time for ``unconditional'' modeling. For inference, we use the Euler method to integrate the velocity vector field $v_t$ using 8 function evaluations (NFEs) and classifier-free guidance (CFG)~\citep{cfg}. Concretely, the form of $v_t$ and $x_t$ is:
\begin{align}
    v_t &= \alpha v_{\theta}(x_t, t, h) + (1 - \alpha)v_{\theta}(x_t, t, \emptyset) \\
    x_{t + \Delta t} &= x_t + v_t \cdot \Delta t
\end{align}
\looseness=-1 where $h$ is the hidden state from decoder backbone and $\emptyset$ is the unconditional case where we pass a vector of zeros with the same shape as $h$. $v_{\theta}(x_t, t, h)$ is the predicted velocity field at time step $t$, sample $x_t$ and conditioning input $h$. We set $\Delta t = 1/8$ and $\alpha = 1.2$ based on the analysis in Section~\ref{sec:cfg_nfe_ablations}.

\looseness=-1 Note that in our architecture, CFG is applied independently at every frame in the FM transformer. Hence, it only requires an extra forward-propagation of only the FM transformer, and is thus significantly cheaper than applying CFG in the decoder backbone. The float values predicted by the FM transformer are converted to discrete integer values by quantizing to the 21 FSQ levels. These discretized tokens are provided as input to the decoder backbone in the next decoding step.

\looseness=-1 Given the inputs to the decoder backbone are discrete tokens with embedding lookup, we also considered alternative architectures inspired by MaskGIT \citep{maskgit} and Depth Transformer \citep{defossez2024moshi}. Both approaches performed reasonably well, but were inferior to FM in human evaluations, especially on expressivity. In addition, MaskGIT requires attending over all 36 acoustic codebook positions and conditioning tokens, resulting in a per-frame sequence length of 38, compared to just 3 in the FM transformer ($h$, $t$, $x_t$). Similarly, the Depth Transformer requires 36 auto-regressive decoding steps, compared to 8 NFEs for FM. Thus, FM is superior in quality, compute and latency.

\section{Training} \label{sec:training}
\subsection{Pretraining}
\looseness=-1 We train the model using paired audio and transcripts pseudo-labelled with Voxtral Mini Transcribe~\citep{voxtral}. Each training sample consists of a tuple $(A_1, T_2, A_2)$ where $A_1$ is a voice reference and $T_2$ is the transcript for $A_2$, which is our target for generation. Similar to Voxtral, we interleave these segments with a \lstinline{<next>} special token between $A_1$ and $T_2$, and a \lstinline{<repeat>} special token between $T_2$ and $A_2$. We ensure that $A_1$ and $A_2$ are single-speaker segments from the same speaker, but not necessarily temporally adjacent. The maximum duration of $A_1$ and $A_2$ are 180 seconds, and we ensure $A_1$ is at least 1 second long. Due to the long-tailed nature of natural conversational human speech duration, we find the model works best on voice prompts ($A_1$) between 3 and 25 seconds.

The loss is computed only on the tokens of $A_2$. We optimize the model using a two-part loss function consisting of a cross-entropy loss on the semantic token $\mathcal{L}_{\text{semantic}}$ and flow-matching loss $\mathcal{L}_{\text{acoustic}}$ on the acoustic tokens. We use the simple conditional flow-matching objective as shown below:

\begin{align}
    \mathcal{L}_{\text{acoustic}} &= \mathbb{E}_{t\sim \mathcal{U}[0, 1],x_0\sim\mathcal{D}, x_1\sim \mathcal{N}(0, 1)}\lVert v_\theta(x_t, t) - u_t(x_t|x_1, x_0)\rVert_2^2 \\
    u_t(x_t|x_1, x_0) &= x_1 - x_0
\end{align}

\looseness=-1 where $u_t$ is the conditional velocity target, $v_\theta$ is the velocity predicted by the FM transformer, $x_1$ is sampled from a normal distribution, and $x_0$ the data distribution $\mathcal{D}$. We initialize the decoder backbone with Ministral 3B. Newly introduced modules, such as the FM transformer, audio codebook embedding lookup-tables and output projection layers, are randomly initialized. During training, we freeze the text-embedding layers in the decoder backbone to improve robustness to text tokens that appear with low-frequency in the Voxtral Mini Transcribe transcriptions. To avoid overfitting to silence, we also use a lower loss weight for frames that have no speech as determined by a voice-activity-detection (VAD) model and set the loss weight to 0 for extremely long silences. We also perform simple LLM based rewrites of the transcripts to introduce robustness to normalized vs un-normalized text (e.g. "5 - 4" vs "five minus four").

\subsection{Direct Preference Optimization}

We use Direct Preference Optimization (DPO)~\citep{dpo} to post-train the model, focusing on improving word error rate (WER) and speaker similarity. For the semantic codebook, we use the standard DPO objective. Given that the acoustic codebooks are predicted with flow-matching, we adapt the objective from~\cite{flow-dpo}:

\begin{equation}
\mathcal{L}(\theta)
=
- \mathbb{E}_{t \sim \mathcal{U}(0,1),\, x^w, x^l}
\log \sigma\!\left(
- \beta \big(
\Delta_\theta(x^w, x^l, t)
- \Delta_{\theta_{\text{ref}}}(x^w, x^l, t)
\big)
\right),
\end{equation}
\noindent
where
\begin{equation}
\Delta_\theta(x^w, x^l, t)
=
\lVert v_\theta(x_t^w, t) - u_t(x_t^w|x^w) \rVert^2_2
-
\lVert v_\theta(x_t^l, t) - u_t(x_t^l|x^l) \rVert^2_2.
\end{equation}

We make the objective suitable for our auto-regressive setup (note the bold $\boldsymbol{t}$ showing each token has a differently sampled t) by computing:
\begin{equation}
\Delta_{\theta}(x^w, x^l, \boldsymbol{t}) = \sum_{i=1} ^{N_w} \lVert v_{\theta}(x_{i,\boldsymbol{t}_i}^w, \boldsymbol{t}_i) - u_{i,\boldsymbol{t}_i}^w \rVert^2_2 - \sum_{i=1} ^{N_l} \lVert v_{\theta}(x_{i,\boldsymbol{t}_i}^l, \boldsymbol{t}_i) - u_{i,\boldsymbol{t}_i}^l \rVert^2_2
\end{equation}
and find that length normalization (dividing by length of winner) causes instability.

We ensure that the $t$ and $x_0$ sampled for each location in the sequence is consistent for the policy model $\theta$ and reference model $\theta_{\text{ref}}$. The two DPO losses are added with uniform weights but we use a $\beta_{\text{semantic}} = 0.1$ and $\beta_{\text{acoustic}} = 0.5$ as training is sensitive to the flow-DPO loss. A low learning rate of $8\mathrm{e}{-8}$ is used for training stability.

The data for DPO is gathered using a rejection-sampling pipeline that takes as input a set of voice samples from a held-out set of single-speaker voice samples and diverse synthetically generated text-prompts. We prompt Mistral Small Creative \footnote{\url{https://docs.mistral.ai/models/mistral-small-creative-25-12}} with the transcript of the voice prompt and randomly chosen personas to synthesize a diverse array of texts which continue or reply to the conversational context. The pretrained checkpoint then takes as input the voice and text prompts and generates multiple samples from each input, from which winner and loser pairs can be constructed. Winners and losers are determined from WER, speaker similarity, loudness consistency, UTMOS-v2~\citep{utmos} and other LM judge metrics. We optimize the model using the combined DPO loss along with the pretraining objective on high-quality speech for 1 epoch, as we found that training longer on synthetic data led to more robotic speech.

\section{Results}
\label{sec:results}
\subsection{Voxtral Codec} \label{sec:codec-results}
Table~\ref{tab:codec-comparison} shows a comparison between Voxtral Codec and Mimi on the Expresso dataset~\citep{nguyen2023expresso}. We evaluate on the following objective metrics: Mel distance, STFT distance, perceptual evaluation of speech quality (PESQ), extended short-time objective intelligibility (ESTOI), word error rate between transcriptions generated using an ASR model corresponding to the source and reconstruction (ASR-WER), speaker similarity score computed using a speaker embedding model. We also report the bitrates and frames per second (fps), which are relevant as these codecs are used in the context of auto-regressive decoder models. Given Mimi uses an RVQ design for acoustic codebooks, it has the flexibility to choose a subset of codebooks to trade-off bitrate and quality. When Voxtral Codec is compared to Mimi in a 16 codebook configuration, such that the bitrates are similar, Voxtral Codec outperforms on all the objective metrics. On an internal subjective assessment, we found Voxtral Codec to be comparable or better than Mimi at 16 codebooks on audios consisting of speech which is our main focus.

\begin{table}[ht]
  \centering
  \caption{Comparison of Voxtral Codec and Mimi on the Expresso dataset.}
  \label{tab:codec-comparison}
  \begin{adjustbox}{max width=\textwidth}
    \begin{tabular}{l c c c c c c c c c}
      \toprule
      \multirow{2}{*}{\textbf{Model}} & \multirow{2}{*}{fps} & \multirow{2}{2.8cm}{\centering token/frame $\times$ vocab. size} & bitrate & \multicolumn{2}{c}{Reconstruction ($\downarrow$)} & \multicolumn{2}{c}{Intrusive ($\uparrow$)} & \multicolumn{2}{c}{Perceptual} \\
      \cmidrule(lr){5-6} \cmidrule(lr){7-8} \cmidrule(lr){9-10}
       & &  & (kbps) & Mel & STFT & PESQ & ESTOI & ASR-WER (\%)$\downarrow$ & Speaker\ Sim$\uparrow$ \\
      \midrule
      Mimi -- 8cb (Moshi)            & 12.5 & 8 $\times$ (2048)     & 1.1  & 0.702 & 1.177 & 2.07 & 0.803 & 11.75 & 0.672 \\
      Mimi -- 16cb                   & 12.5 & 16 $\times$ (2048)    & 2.2  & 0.618 & 1.100 & 2.67 & 0.865 & 11.01 & 0.829 \\
      Mimi -- full 32cb              & 12.5 & 32 $\times$ (2048)    & 4.4  & 0.552 & 1.040 & 3.18 & 0.910 & 10.25 & 0.902 \\
      \midrule
      Voxtral Codec                  & 12.5 & 1 $\times$ (8192) + 36 $\times$ (21) & 2.1  & 0.545 & 0.982 & 3.05 & 0.882 & 10.66 & 0.843 \\
      \bottomrule
    \end{tabular}
  \end{adjustbox}
\end{table}

\subsection{Automatic Evaluations}

We evaluate Voxtral TTS, ElevenLabs v3 and ElevenLabs Flash v2.5 on SEED-TTS~\citep{seedtts} and the nine supported languages in MiniMax-TTS~\citep{minimaxtts} using automated metrics:
\begin{enumerate}
    \item \textbf{Word Error Rate (WER)}: Measured by Voxtral Mini Transcribe v2 to capture the intelligibility of speech.
    \item \textbf{UTMOS-v2}~\citep{utmos}: Predicts the Mean Opinion Score (MOS) of generated speech.
    \item \textbf{Speaker Similarity}: Speaker embeddings are predicted using the ECAPA-TDNN model~\citep{ecapa} and the cosine similarity is computed against the reference embedding. This evaluates how closely generated speech emulates the provided voice reference.
\end{enumerate}

The results for the three models are presented in Table~\ref{tab:main_results_automated}. While both ElevenLabs models achieve low WERs across languages, Voxtral TTS significantly outperforms ElevenLabs on the speaker similarity metrics. Surprisingly, we find that ElevenLabs Flash v2.5 performs better on most automated metrics and ElevenLabs v3 better on human evaluations, particularly with emotion steering. This highlights the importance of performing human evaluations in conjunction with automatic evaluations.

\begin{table}[h] 
  \centering
  \caption{
  WER, UTMOS, and Speaker Similarity scores for Voxtral TTS, ElevenLabs v3, and ElevenLabs Flash v2.5.}
  \resizebox{\textwidth}{!}{
  \begin{tabular}{lccccccccc}
  \toprule
  & \multicolumn{3}{c}{WER (\%) $\downarrow$} & \multicolumn{3}{c}{UTMOS $\uparrow$} & \multicolumn{3}{c}{Speaker Sim
  $\uparrow$} \\
  \cmidrule(lr){2-4} \cmidrule(lr){5-7} \cmidrule(lr){8-10}
  Task & Voxtral & ElevenLabs v3 & ElevenLabs Flash & Voxtral & ElevenLabs v3 & ElevenLabs Flash & Voxtral & ElevenLabs v3 & ElevenLabs Flash
  \\
  \midrule
  \multicolumn{10}{l}{\textit{MiniMax}} \\
  Arabic & \textbf{2.68} & 3.67 & 2.86 & \textbf{3.07} & 2.50 & 2.89 & \textbf{0.746} & 0.546 & 0.539 \\
  German & 0.83 & \textbf{0.45} & 1.08 & 3.12 & 2.90 & \textbf{3.27} & \textbf{0.721} & 0.457 & 0.489 \\
  English & 0.63 & 0.48 & \textbf{0.33} & \textbf{4.30} & 4.27 & 4.27 & \textbf{0.786} & 0.484 & 0.489 \\
  Spanish & 0.51 & 0.87 & \textbf{0.49} & \textbf{3.41} & 3.18 & 2.99 & \textbf{0.762} & 0.443 & 0.541 \\
  French & 3.22 & 2.34 & \textbf{2.26} & 2.83 & 2.90 & \textbf{2.94} & \textbf{0.587} & 0.339 & 0.378 \\
  Hindi & \textbf{4.99} & 8.71 & 5.08 & \textbf{3.56} & \textbf{3.56} & 3.35 & \textbf{0.839} & 0.707 &  0.679 \\
  Italian & 1.32 & 0.58 & \textbf{0.55} & \textbf{3.43} & 3.08 & 3.09 & \textbf{0.739} & 0.527 & 0.485 \\
  Dutch & 1.99 & 1.52 & \textbf{0.83} & \textbf{3.89} & 3.53 & 3.68 & \textbf{0.720} & 0.397 & 0.598 \\
  Portuguese & 1.02 & \textbf{0.92} & 1.15 & \textbf{3.66} & 3.41 & 3.41 & \textbf{0.785} & 0.571 & 0.642 \\
  \midrule
  Seed TTS & 1.23 & 1.26 & \textbf{0.86} & \textbf{4.11} & 3.92 & 4.09 & \textbf{0.628} & 0.392 & 0.413 \\
  \bottomrule
  \end{tabular}
  }
  \label{tab:main_results_automated}
  \end{table}

\subsection{Human Evaluations}\label{sec:humanevals}
Automated metrics cannot measure the naturalness and expressivity of a TTS model, especially the ability of the model to speak with a specific emotion. We find that UTMOS is only a loose proxy, not well calibrated across languages and only weakly correlated with human preference. Hence, we perform two sets of human evaluations in which annotators compare generations between two models without knowing their identities. The evaluation consists of 77 prompts, with 11 of them neutral while 66 of them have an associated expected emotion. For all evaluations, annotators are instructed to choose whether one of the generations is "slightly better", "much better" or if they are "both good" or "both bad". During labeling, all audio samples are resampled to 24 kHz WAV format (even the reference samples) to ensure there is no bias due to audio quality.

\subsubsection{Flagship voices}
\looseness=-1 First, we compare our flagship voices (\textit{British-Female, British-Male, American-Male, French-Female}) against the flagship voices of same gender and accent provided by competitors. We run two sub-evaluations:
\begin{enumerate}
    \item \textbf{Explicit steering}: We test the ability to bias a TTS model's generation toward a specific emotion. The TTS prompts which have an associated emotion (not Neutral) are provided as free-form instruction to Gemini 2.5 Flash TTS as it supports free-form instructions such as "Speak in an angry tone.". For ElevenLabs v3 we provide emotion tags enclosed in brackets \footnote{\url{https://elevenlabs.io/blog/eleven-v3-audio-tags-expressing-emotional-context-in-speech}}. While Voxtral TTS does not support emotion tags/text-instructions, we steer the generation by leveraging a different voice prompt provided from the same speaker which embodies the requested emotion.
    \item \textbf{Implicit steering}: We test the model's capabilities to infer emotion from provided text (e.g. "This is the best day of my life!"). No emotion label or instruction is provided to the model. For Voxtral TTS, we use a neutral voice prompt.
\end{enumerate}

We use three annotators who are native speakers of the same dialect for each language per pair. The win rates of Voxtral TTS (excluding ties) are presented in Table~\ref{tab:model_winrate}. Gemini 2.5 Flash TTS is the strongest model, and Voxtral TTS is competitive against ElevenLabs v3. In the implicit steering setting, Voxtral TTS consistently outperforms both ElevenLabs models.

\begin{table}[h]
\centering
\caption{
\textbf{Voxtral TTS win rates by steering type.} In the explicit steering setting, Voxtral TTS is competitive with ElevenLabs v3, while having a higher win rate compared to both ElevenLabs models in the implicit steering setting.}
\begin{tabular}{llc}
\toprule
\textbf{Emotion steering} & \textbf{Opponent Model} & \textbf{Voxtral TTS Win Rate (\%)} \\ \midrule
\multirow{2}{*}{Explicit} & ElevenLabs v3         & 51.0 \\
                                   & Gemini 2.5 Flash TTS & 35.4 \\ \midrule
\multirow{2}{*}{Implicit} & ElevenLabs Flash v2.5 & 58.3 \\
                                   & ElevenLabs v3         & 55.4 \\
                                   & Gemini 2.5 Flash TTS         & 37.1 \\
\bottomrule
\end{tabular}
\label{tab:model_winrate}
\end{table}

\subsubsection{Zero-Shot Voice Cloning}

To evaluate voice cloning capabilities, we source high quality audios from two recognized speakers in each language. We generate speech from each model in a zero-shot setting, and instruct annotators to rate the generations based on (a) likeness of the generated audio to voice prompt and (b) naturalness of speech and expressivity.


\begin{table}[h]
\centering
\caption{\textbf{Voxtral TTS win rate against ElevenLabs Flash v2.5 across languages.} Voxtral TTS matches or outperforms ElevenLabs Flash v2.5 on every language, and has an overall micro-average win rate of 68.4\%.}
\label{tab:win-rates}
\begin{tabular}{lc}
\toprule
\textbf{Language} & \textbf{Voxtral TTS Win Rate} (\%) \\
\midrule
Arabic    & 72.9 \\
Dutch     & 49.4 \\
English   & 60.8 \\
French    & 54.4 \\
German    & 72.0 \\
Hindi     & 79.8 \\
Italian   & 57.1 \\
Portuguese & 74.4 \\
Spanish   & 87.8 \\
\midrule
\textbf{Overall} & \textbf{68.4} \\
\bottomrule
\end{tabular}

\end{table}

Table~\ref{tab:win-rates} shows the Voxtral TTS win rates against ElevenLabs Flash v2.5 across languages. Overall, Voxtral TTS has a win rate of 68.4\%, with significantly better results across both high and low-resource languages (such as Arabic and Hindi). Notably, the Voxtral TTS win rate is much higher in the zero-shot setting (68.4\%) compared with flagship voices (58.3\%), highlighting that Voxtral TTS is a far more generalizable model, capturing a diverse range of user-voices.

\section{Analysis}
\label{sec:analysis}
In this Section, we provide a comparison between the pretrained and DPO checkpoints and ablate the pertinent inference parameters.

\subsection{DPO improvements} \label{sec:dpo_improvements}

Table~\ref{tab:pt-vs-dpo} shows the WER and UTMOS metrics for the pretrained and DPO checkpoints. Overall, DPO improves on both metrics, with the largest gains in German and French and regressions only on Hindi. Qualitatively, we find that the DPO model hallucinates less and skips fewer words. DPO also ameliorates the pretrained model's occasional tendency to significantly taper in volume throughout the audio. Interestingly, DPO has minimal effect on speaker similarity, which is within $\pm0.01$ of the pretrained checkpoint (not presented here for brevity).

\begin{table}[h]
\centering
\caption{DPO improves WER and UTMOS across languages.}
\label{tab:pt-vs-dpo}
\begin{tabular}{lrrrr}
\toprule
& \multicolumn{2}{c}{WER (\%) $\downarrow$} & \multicolumn{2}{c}{UTMOS $\uparrow$} \\
\cmidrule(lr){2-3} \cmidrule(lr){4-5}
Task & Pretrain & DPO & Pretrain & DPO \\
\midrule
\multicolumn{5}{l}{\textit{MiniMax}} \\
Arabic & 2.80 & \textbf{2.68} \small{(-0.12)} & 3.01 & \textbf{3.07} \small{(+0.06)} \\
German & 4.08 & \textbf{0.83} \small{(-3.25)} & 3.05 & \textbf{3.12} \small{(+0.07)} \\
English & 0.84 & \textbf{0.63} \small{(-0.21)} & 4.25 & \textbf{4.30} \small{(+0.05)} \\
Spanish & 0.56 & \textbf{0.51} \small{(-0.06)} & 3.38 & \textbf{3.41} \small{(+0.04)} \\
French & 5.01 & \textbf{3.22} \small{(-1.79)} & 2.76 & \textbf{2.83} \small{(+0.07)} \\
Hindi & \textbf{3.39} & 4.99 \small{(+1.61)} & 3.43 & \textbf{3.56} \small{(+0.13)} \\
Italian & 2.18 & \textbf{1.32} \small{(-0.85)} & 3.36 & \textbf{3.43} \small{(+0.07)} \\
Dutch & 3.10 & \textbf{1.99} \small{(-1.11)} & 3.85 & \textbf{3.89} \small{(+0.04)} \\
Portuguese & 1.17 & \textbf{1.02} \small{(-0.15)} & 3.60 & \textbf{3.66} \small{(+0.06)} \\
\midrule
Seed TTS & 1.58 & \textbf{1.23} \small{(-0.35)} & 4.07 & \textbf{4.11} \small{(+0.04)} \\
\bottomrule
\end{tabular}
\end{table}

\subsection{Inference Parameters} \label{sec:cfg_nfe_ablations}

\begin{center}
\begin{figure}[h!]
    \includegraphics[width=0.95\linewidth, trim={5 5 5 5}, clip]{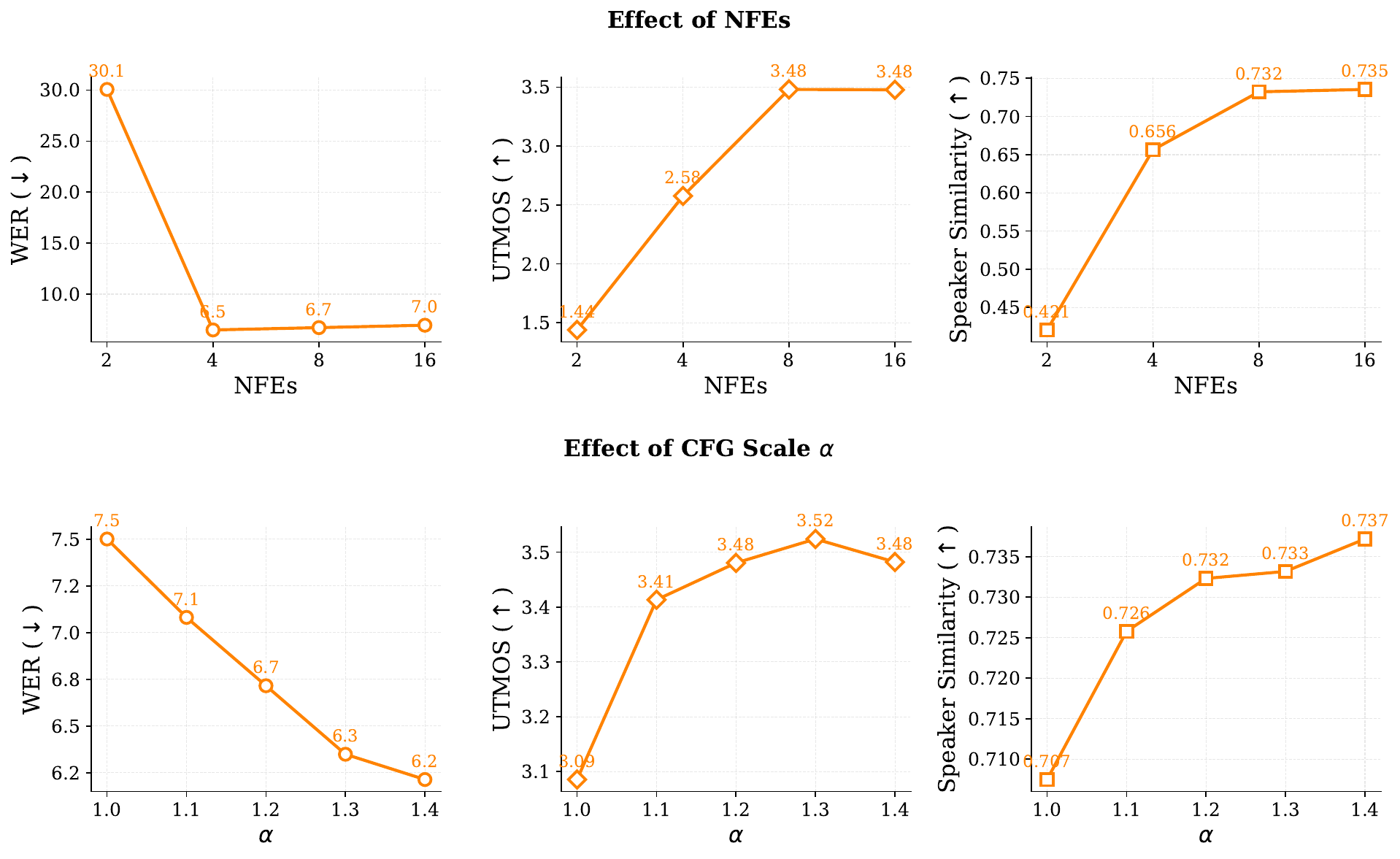}
    \caption{
    \textbf{Effect of NFEs and CFG on automatic evaluations.} The metrics are averaged over SEED-TTS and the 9 languages in MiniMax. Increasing the NFEs from 2 to 8 improves speaker similarity and UTMOS metrics. There is a slight regression in WER as the NFEs is increased beyond this. The metrics monotonically increase with higher CFG, but human evaluations flagged regressions in text-adherence with high $\alpha$.}
    \label{fig:abl_nfe_cfg}
\end{figure}
\end{center}

Figure~\ref{fig:abl_nfe_cfg} demonstrates the effect on the automatic evaluation metrics as the number of functional evaluations (NFEs) and choice of CFG $\alpha$ are varied. There are marked improvements across metrics as the NFEs is increased from 2 to 8. We find that increasing number of NFEs beyond 8 yields marginal improvement in speaker similarity and minor degradations in WER. Thus, we chose 8 NFEs as our default inference setting.

Increasing the value of CFG $\alpha$, we find that there is a nearly monotonic improvement in all metrics except UTMOS-v2. However, internal human evaluations found that a higher $\alpha$ led to over-adherence to the provided voice-prompt and the model failed to bias towards emotions that are implicit in the text prompt. We also find that lower $\alpha = 1.2$ works best for higher quality audio (e.g. professional recordings), while in-the-wild recordings might benefit from a higher $\alpha$.

\section{Inference and Serving in vLLM-Omni}
\label{sec:inference}
Voxtral TTS is served through vLLM-Omni~\citep{vllm-omni}, an extension of the vLLM~\citep{vllm} for multi-stage multimodal models. Voxtral TTS is decomposed into a two-stage pipeline: a generation stage that predicts the audio tokens (semantic and acoustic), followed by a codec decoding stage that converts the tokens into a waveform. The two stages communicate through an asynchronous chunked streaming protocol over shared memory, enabling first-audio latency well before the full waveform has been generated.

\subsection{CUDA Graph Acceleration for Flow-Matching Transformer}
The flow-matching transformer is the computational bottleneck of the generation stage. Each decoding step requires $N$ function evaluations with CFG, requiring $2 \times N$ forward passes per generated frame.

To eliminate Python-level overhead and kernel-launch latency, the entire ODE solver is captured into CUDA graphs. At startup, an eager warmup pass is performed for each bucket size and the corresponding CUDA graph is then captured. During inference, the actual batch size is rounded up to the nearest bucket by padding the input with zeros. Next, the CUDA graph is replayed and outputs are sliced back to the actual batch size. If the batch size exceeds the largest captured bucket, the model falls back to eager execution.

To evaluate the effect of CUDA graph acceleration, we compare the latency and real-time factor (RTF) when decoding with eager mode and CUDA graphs. Table~\ref{tab:cudagraph-ablation} reports results for a 500-character text input, a 10-second audio reference and concurrency 1 on a single H200. Enabling CUDA graphs results in a 47\% improvement to latency and reduces the RTF by 2.5x.

\begin{table}[h]
\centering
\caption{Effect of CUDA graph acceleration on the flow-matching transformer.}
\label{tab:cudagraph-ablation}
\begin{tabular}{lcccc}
\toprule
\textbf{Configuration}  & \textbf{Latency} & \textbf{RTF} \\
\midrule
  Eager mode & 133 ms & 0.258 \\
  CUDA graph   &  70 ms      &  0.103 \\
\bottomrule
\end{tabular}
\end{table}


\subsection{Asynchronous Chunked Streaming}
The two pipeline stages run in separate scheduling loops. To overlap the autoregressive generation stage decoding with codec decoding stage waveform synthesis, an asynchronous chunked streaming protocol is introduced.

After each generation step, the vLLM-Omni transfer manager stores the audio tokens to a per-request buffer. Once the buffer length reaches a pre-defined length, a chunk of tokens are emitted to the codec decoding stage. To ensure coherence between chunks, each emitted chunk includes a slice of previous frames in addition to the new frames. This overlap enables the codec decoder’s causal sliding-window attention to maintain temporal coherence across chunk boundaries.

\subsection{Inference Throughput}

With the techniques introduced in this section, Voxtral TTS achieves low-latency, high-throughput inference. Table~\ref{tab:serving-perf} shows the serving performance on a single NVIDIA H200 from concurrency 1 to 32 with 500-character text inputs and 10-second audio references. As the concurrency is increased from 1 to 32, the throughput scales from 119 to 1{,}431 characters per second per GPU, a 12x increase, while latency remains sub-second. The wait rate, defined as the fraction of audio chunks for which the client must stall since it is waiting for outputs, remains zero across all concurrency levels. As concurrency grows, per-request RTF increases modestly to 0.302 at concurrency 32, still well within the real-time boundary. 

\looseness=-1 These results demonstrate that Voxtral TTS is suitable for production deployment: a single H200 can serve over 30 concurrent users with uninterrupted streaming output and sub-second time to first audio.

\begin{table}[h]
\centering
\caption{Serving performance of Voxtral TTS on a single H200.}
\label{tab:serving-perf}
\begin{tabular}{ccccc}
  \toprule
  \textbf{Concurrency} & \textbf{Latency} & \textbf{RTF} & \textbf{Throughput (char/s/GPU)} & \textbf{Wait Rate} \\
  \midrule
  1  & 70 ms & 0.103 & 119.14 & 0\% \\
  16 & 331 ms & 0.237 & 879.11 & 0\% \\
  32 & 552 ms & 0.302 & 1430.78 & 0\% \\
  \bottomrule
\end{tabular}
\end{table}



\section{Conclusion}
\label{sec:conclusion}
We introduced Voxtral TTS, a multilingual TTS model that leverages a hybrid architecture for auto-regressive generation of semantic tokens and flow-matching for acoustic tokens. The tokens correspond to those from Voxtral Codec, a speech tokenizer that combines ASR-distilled semantic tokens with FSQ acoustic tokens.

\looseness=-1 Voxtral TTS is able to generate expressive, voice-cloned speech from as little as 3 seconds of reference audio, and is preferred to API baselines in human evaluations. We release Voxtral TTS as open weights under the CC BY-NC license to support further research and development of expressive TTS systems.

\subsection*{Core contributors}

Alexander H. Liu, Alexis Tacnet, Andy Ehrenberg, Andy Lo, Chen-Yo Sun, Guillaume Lample, Henry Lagarde, Jean-Malo Delignon, Jaeyoung Kim, John Harvill, Khyathi Raghavi Chandu, Lorenzo Signoretti, Margaret Jennings, Patrick von Platen, Pavankumar Reddy Muddireddy, Rohin Arora, Sanchit Gandhi, Samuel Humeau, Soham Ghosh, Srijan Mishra, Van Phung.

\subsection*{Contributors}

Abdelaziz Bounhar, Abhinav Rastogi, Adrien Sadé, Alan Jeffares, Albert Jiang, Alexandre Cahill, Alexandre Gavaudan, Alexandre Sablayrolles, Amélie Héliou, Amos You, Andrew Bai, Andrew Zhao, Angele Lenglemetz, Anmol Agarwal, Anton Eliseev, Antonia Calvi, Arjun Majumdar, Arthur Fournier, Artjom Joosen, Avi Sooriyarachchi, Aysenur Karaduman Utkur, Baptiste Bout, Baptiste Rozière, Baudouin De Monicault, Benjamin Tibi, Bowen Yang, Charlotte Cronjäger, Clémence Lanfranchi, Connor Chen, Corentin Barreau, Corentin Sautier, Cyprien Courtot, Darius Dabert, Diego de las Casas, Elizaveta Demyanenko, Elliot Chane-Sane, Emmanuel Gottlob, Enguerrand Paquin, Etienne Goffinet, Fabien Niel, Faruk Ahmed, Federico Baldassarre, Gabrielle Berrada, Gaëtan Ecrepont, Gauthier Guinet, Genevieve Hayes, Georgii Novikov, Giada Pistilli, Guillaume Kunsch, Guillaume Martin, Guillaume Raille, Gunjan Dhanuka, Gunshi Gupta, Han Zhou, Harshil Shah, Hope McGovern, Hugo Thimonier, Indraneel Mukherjee, Irene Zhang, Jacques Sun, Jan Ludziejewski, Jason Rute, Jérémie Dentan, Joachim Studnia, Jonas Amar, Joséphine Delas, Josselin Somerville Roberts, Julien Tauran, Karmesh Yadav, Kartik Khandelwal, Kilian Tep, Kush Jain, Laurence Aitchison, Laurent Fainsin, Léonard Blier, Lingxiao Zhao, Louis Martin, Lucile Saulnier, Luyu Gao, Maarten Buyl, Manan Sharma, Marie Pellat, Mark Prins, Martin Alexandre, Mathieu Poirée, Mathieu Schmitt, Mathilde Guillaumin, Matthieu Dinot, Matthieu Futeral, Maxime Darrin, Maximilian Augustin, Mert Unsal, Mia Chiquier, Mikhail Biriuchinskii, Minh-Quang Pham, Mircea Lica, Morgane Rivière, Nathan Grinsztajn, Neha Gupta, Olivier Bousquet, Olivier Duchenne, Patricia Wang, Paul Jacob, Paul Wambergue, Paula Kurylowicz, Philippe Pinel, Philomène Chagniot, Pierre Stock, Piotr Miłoś, Prateek Gupta, Pravesh Agrawal, Quentin Torroba, Ram Ramrakhya, Randall Isenhour, Rishi Shah, Romain Sauvestre, Roman Soletskyi, Rosalie Millner, Rupert Menneer, Sagar Vaze, Samuel Barry, Samuel Belkadi, Sandeep Subramanian, Sean Cha, Shashwat Verma, Siddhant Waghjale, Siddharth Gandhi, Simon Lepage, Sumukh Aithal, Szymon Antoniak, Tarun Kumar Vangani, Teven Le Scao, Théo Cachet, Theo Simon Sorg, Thibaut Lavril, Thomas Chabal, Thomas Foubert, Thomas Robert, Thomas Wang, Tim Lawson, Tom Bewley, Tom Edwards, Tyler Wang, Umar Jamil, Umberto Tomasini, Valeriia Nemychnikova, Vedant Nanda, Victor Jouault, Vincent Maladière, Vincent Pfister, Virgile Richard, Vladislav Bataev, Wassim Bouaziz, Wen-Ding Li, William Havard, William Marshall, Xinghui Li, Xingran Guo, Xinyu Yang, Yannic Neuhaus, Yassine El Ouahidi, Yassir Bendou, Yihan Wang, Yimu Pan, Zaccharie Ramzi, Zhenlin Xu.

\subsection{Acknowledgements}

We would like to thank Han Gao, Hongsheng Liu, Roger Wang, and Yueqian Lin from the vLLM-Omni team for their support and contributions in integrating Voxtral TTS into the vLLM-Omni framework.

\clearpage

\bibliography{ref}

@article{defossez2022high,
  title={High fidelity neural audio compression},
  author={D{\'e}fossez, Alexandre and Copet, Jade and Synnaeve, Gabriel and Adi, Yossi},
  journal={arXiv preprint arXiv:2210.13438},
  year={2022}
}

@article{parker2024scaling,
  title={Scaling transformers for low-bitrate high-quality speech coding},
  author={Parker, Julian D and Smirnov, Anton and Pons, Jordi and Carr, CJ and Zukowski, Zack and Evans, Zach and Liu, Xubo},
  journal={arXiv preprint arXiv:2411.19842},
  year={2024}
}

@article{wu2024ts3,
  title={Ts3-codec: Transformer-based simple streaming single codec},
  author={Wu, Haibin and Kanda, Naoyuki and Eskimez, Sefik Emre and Li, Jinyu},
  journal={arXiv preprint arXiv:2411.18803},
  year={2024}
}

@article{press2021train,
  title={Train short, test long: Attention with linear biases enables input length extrapolation},
  author={Press, Ofir and Smith, Noah A and Lewis, Mike},
  journal={arXiv preprint arXiv:2108.12409},
  year={2021}
}

@inproceedings{touvron2021going,
  title={Going deeper with image transformers},
  author={Touvron, Hugo and Cord, Matthieu and Sablayrolles, Alexandre and Synnaeve, Gabriel and J{\'e}gou, Herv{\'e}},
  booktitle={Proceedings of the IEEE/CVF international conference on computer vision},
  pages={32--42},
  year={2021}
}

@article{van2017neural,
  title={Neural discrete representation learning},
  author={Van Den Oord, Aaron and Vinyals, Oriol and others},
  journal={Advances in neural information processing systems},
  volume={30},
  year={2017}
}

@article{zhang2023speechtokenizer,
  title={Speechtokenizer: Unified speech tokenizer for speech large language models},
  author={Zhang, Xin and Zhang, Dong and Li, Shimin and Zhou, Yaqian and Qiu, Xipeng},
  journal={arXiv preprint arXiv:2308.16692},
  year={2023}
}

@article{defossez2024moshi,
  title={Moshi: a speech-text foundation model for real-time dialogue},
  author={D{\'e}fossez, Alexandre and Mazar{\'e}, Laurent and Orsini, Manu and Royer, Am{\'e}lie and P{\'e}rez, Patrick and J{\'e}gou, Herv{\'e} and Grave, Edouard and Zeghidour, Neil},
  journal={arXiv preprint arXiv:2410.00037},
  year={2024}
}

@inproceedings{liu2024revisiting,
  title={Revisiting self-supervised learning of speech representation from a mutual information perspective},
  author={Liu, Alexander H and Yeh, Sung-Lin and Glass, James R},
  booktitle={ICASSP 2024-2024 IEEE International Conference on Acoustics, Speech and Signal Processing (ICASSP)},
  pages={12051--12055},
  year={2024},
  organization={IEEE}
}

@article{vashishth2024stab,
  title={STAB: Speech tokenizer assessment benchmark},
  author={Vashishth, Shikhar and Singh, Harman and Bharadwaj, Shikhar and Ganapathy, Sriram and Asawaroengchai, Chulayuth and Audhkhasi, Kartik and Rosenberg, Andrew and Bapna, Ankur and Ramabhadran, Bhuvana},
  journal={arXiv preprint arXiv:2409.02384},
  year={2024}
}

@inproceedings{radford2023robust,
  title={Robust speech recognition via large-scale weak supervision},
  author={Radford, Alec and Kim, Jong Wook and Xu, Tao and Brockman, Greg and McLeavey, Christine and Sutskever, Ilya},
  booktitle={International conference on machine learning},
  pages={28492--28518},
  year={2023},
  organization={PMLR}
}

@article{liu2026ministral,
  title={Ministral 3},
  author={Liu, Alexander H and Khandelwal, Kartik and Subramanian, Sandeep and Jouault, Victor and Rastogi, Abhinav and Sad{\'e}, Adrien and Jeffares, Alan and Jiang, Albert and Cahill, Alexandre and Gavaudan, Alexandre and others},
  journal={arXiv preprint arXiv:2601.08584},
  year={2026}
}

@inproceedings{berndt1994dtw,
author = {Berndt, Donald J. and Clifford, James},
title = {Using dynamic time warping to find patterns in time series},
year = {1994},
publisher = {AAAI Press},
booktitle = {Proceedings of the 3rd International Conference on Knowledge Discovery and Data Mining},
pages = {359–370},
numpages = {12},
location = {Seattle, WA},
series = {AAAIWS'94}
}

@InProceedings{gradtts,
  title     = {Grad-{TTS}: A Diffusion Probabilistic Model for Text-to-Speech},
  author    = {Popov, Vadim and Vovk, Ivan and Gogoryan, Vladimir and Sadekova, Tasnima and Kudinov, Mikhail},
  booktitle = {Proceedings of the 38th International Conference on Machine Learning},
  pages     = {8599--8608},
  year      = {2021},
  volume    = {139},
  series    = {Proceedings of Machine Learning Research},
  publisher = {PMLR},
  url       = {https://proceedings.mlr.press/v139/popov21a.html}
}

@article{audiolm,
  title   = {AudioLM: A Language Modeling Approach to Audio Generation},
  author  = {Borsos, Zal{\'a}n and Marinier, Rapha{\"e}l and Vincent, Damien and
             Kharitonov, Eugene and Pietquin, Olivier and Sharifi, Matt and
             Roblek, Dominik and Teboul, Olivier and Grangier, David and
             Tagliasacchi, Marco and Zeghidour, Neil},
  journal = {IEEE/ACM Transactions on Audio, Speech, and Language Processing},
  volume  = {31},
  pages   = {2523--2533},
  year    = {2023},
  doi     = {10.1109/TASLP.2023.3288409}
}

@article{vall_e,
  title         = {Neural Codec Language Models are Zero-Shot Text to Speech Synthesizers},
  author        = {Chengyi Wang and Sanyuan Chen and Yu Wu and Ziqiang Zhang and Long Zhou and Shujie Liu and
                   Zhuo Chen and Yanqing Liu and Huaming Wang and Jinyu Li and Lei He and Sheng Zhao and Furu Wei},
  journal       = {arXiv preprint arXiv:2301.02111},
  year          = {2023},
  archivePrefix = {arXiv},
  eprint        = {2301.02111},
  url           = {https://arxiv.org/abs/2301.02111}
}

@inproceedings{voicebox,
  title     = {Voicebox: Text-Guided Multilingual Universal Speech Generation at Scale},
  author    = {Le, Matt and Vyas, Apoorv and Shi, Bowen and Karrer, Brian and
               Sari, Leda and Moritz, Rashel and Williamson, Mary and
               Manohar, Vimal and Adi, Yossi and Mahadeokar, Jay and Hsu, Wei-Ning},
  booktitle = {Advances in Neural Information Processing Systems},
  volume    = {36},
  year      = {2023},
  url       = {https://proceedings.neurips.cc/paper_files/paper/2023/hash/2d8911db9ecedf866015091b28946e15-Abstract-Conference.html}
}

@inproceedings{ecapa,
  title     = {{ECAPA-TDNN}: Emphasized Channel Attention, Propagation and Aggregation in {TDNN} Based Speaker Verification},
  author    = {Brecht Desplanques and Jenthe Thienpondt and Kris Demuynck},
  booktitle = {Interspeech 2020},
  pages     = {3830--3834},
  year      = {2020},
  doi       = {10.21437/Interspeech.2020-2650}
}

@misc{vllm-omni,
  title         = {vLLM-Omni: Fully Disaggregated Serving for Any-to-Any Multimodal Models},
  author        = {Peiqi Yin and Jiangyun Zhu and Han Gao and Chenguang Zheng and
                   Yongxiang Huang and Taichang Zhou and Ruirui Yang and Weizhi Liu and
                   Weiqing Chen and Canlin Guo and Didan Deng and Zifeng Mo and
                   Cong Wang and James Cheng and Roger Wang and Hongsheng Liu},
  year          = {2026},
  eprint        = {2602.02204},
  archivePrefix = {arXiv},
  primaryClass  = {cs.DC},
  url           = {https://arxiv.org/abs/2602.02204}
}

@inproceedings{vllm,
  title     = {Efficient Memory Management for Large Language Model Serving with {PagedAttention}},
  author    = {Woosuk Kwon and Zhuohan Li and Siyuan Zhuang and Ying Sheng and
               Lianmin Zheng and Cody Hao Yu and Joseph E. Gonzalez and
               Hao Zhang and Ion Stoica},
  booktitle = {Proceedings of the ACM SIGOPS 29th Symposium on Operating Systems Principles},
  year      = {2023},
  doi       = {10.1145/3600006.3613165}
}

@misc{flow-dpo,
  title         = {MR-FlowDPO: Multi-Reward Direct Preference Optimization for Flow-Matching Text-to-Music Generation},
  author        = {Alon Ziv and Sanyuan Chen and Andros Tjandra and Yossi Adi and Wei-Ning Hsu and Bowen Shi},
  year          = {2025},
  eprint        = {2512.10264},
  archivePrefix = {arXiv},
  primaryClass  = {cs.SD},
  url           = {https://arxiv.org/abs/2512.10264}
}

@inproceedings{dpo,
  title     = {Direct Preference Optimization: Your Language Model is Secretly a Reward Model},
  author    = {Rafael Rafailov and Archit Sharma and Eric Mitchell and Stefano Ermon and Christopher D. Manning and Chelsea Finn},
  booktitle = {Advances in Neural Information Processing Systems},
  year      = {2023},
  url       = {https://arxiv.org/abs/2305.18290}
}

@misc{fsq,
  title         = {Finite Scalar Quantization: {VQ-VAE} Made Simple},
  author        = {Fabian Mentzer and David Minnen and Eirikur Agustsson and Michael Tschannen},
  year          = {2023},
  eprint        = {2309.15505},
  archivePrefix = {arXiv},
  primaryClass  = {cs.CV}
}

@InProceedings{maskgit,
  author    = {Chang, Huiwen and Zhang, Han and Jiang, Lu and Liu, Ce and Freeman, William T.},
  title     = {MaskGIT: Masked Generative Image Transformer},
  booktitle = {Proceedings of the IEEE/CVF Conference on Computer Vision and Pattern Recognition (CVPR)},
  month     = {June},
  year      = {2022},
  pages     = {11315--11325}
}

@InProceedings{dit,
  author    = {Peebles, William and Xie, Saining},
  title     = {Scalable Diffusion Models with Transformers},
  booktitle = {Proceedings of the IEEE/CVF International Conference on Computer Vision (ICCV)},
  month     = {October},
  year      = {2023},
  pages     = {4195--4205}
}

@misc{voxtral,
  title         = {Voxtral},
  author        = {Alexander H. Liu and Andy Ehrenberg and Andy Lo and Cl{\'{e}}ment Denoix and Corentin Barreau and
                   Guillaume Lample and Jean-Malo Delignon and Khyathi Raghavi Chandu and Patrick von Platen and
                   Pavankumar Reddy Muddireddy and Sanchit Gandhi and Soham Ghosh and Srijan Mishra and Thomas Foubert},
  year          = {2025},
  eprint        = {2507.13264},
  archivePrefix = {arXiv},
  primaryClass  = {cs.SD},
  url           = {https://arxiv.org/abs/2507.13264}
}

@inproceedings{utmos,
  title     = {The T05 System for The {VoiceMOS} {Challenge} 2024: Transfer Learning from Deep Image Classifier to Naturalness {MOS} Prediction of High-Quality Synthetic Speech},
  author    = {Baba, Kaito and Nakata, Wataru and Saito, Yuki and Saruwatari, Hiroshi},
  booktitle = {IEEE Spoken Language Technology Workshop (SLT)},
  year      = {2024},
  pages     = {818--824},
  doi       = {10.1109/SLT61566.2024.10832315}
}

@misc{seedtts,
  title         = {Seed-TTS: A Family of High-Quality Versatile Speech Generation Models},
  author        = {Philip Anastassiou and Jiawei Chen and Jitong Chen and Yuanzhe Chen and
                   Zhuo Chen and Ziyi Chen and Jian Cong and Lelai Deng and Chuang Ding and
                   Lu Gao and Mingqing Gong and Peisong Huang and Qingqing Huang and
                   Zhiying Huang and Yuanyuan Huo and Dongya Jia and Chumin Li and Feiya Li and
                   Hui Li and Jiaxin Li and Xiaoyang Li and Xingxing Li and Lin Liu and
                   Shouda Liu and Sichao Liu and Xudong Liu and Yuchen Liu and Zhengxi Liu and
                   Lu Lu and Junjie Pan and Xin Wang and Yuping Wang and Yuxuan Wang and
                   Zhengnan Wei and Jian Wu and Chao Yao and Yifeng Yang and Yuanhao Yi and
                   Junteng Zhang and Qidi Zhang and Shuo Zhang and Wenjie Zhang and
                   Yang Zhang and Zilin Zhao and Dejian Zhong and Xiaobin Zhuang},
  year          = {2024},
  eprint        = {2406.02430},
  archivePrefix = {arXiv},
  primaryClass  = {eess.AS},
  url           = {https://arxiv.org/abs/2406.02430}
}

@misc{minimaxtts,
  title         = {MiniMax-Speech: Intrinsic Zero-Shot Text-to-Speech with a Learnable Speaker Encoder},
  author        = {Bowen Zhang and Congchao Guo and Geng Yang and Hang Yu and Haozhe Zhang and
                   Heidi Lei and Jialong Mai and Junjie Yan and Kaiyue Yang and Mingqi Yang and
                   Peikai Huang and Ruiyang Jin and Sitan Jiang and Weihua Cheng and Yawei Li and
                   Yichen Xiao and Yiying Zhou and Yongmao Zhang and Yuan Lu and Yucen He},
  year          = {2025},
  eprint        = {2505.07916},
  archivePrefix = {arXiv},
  primaryClass  = {eess.AS},
  url           = {https://arxiv.org/abs/2505.07916}
}

@article{cfg,
  title   = {Classifier-Free Diffusion Guidance},
  author  = {Jonathan Ho and Tim Salimans},
  journal = {arXiv preprint arXiv:2207.12598},
  year    = {2022},
  url     = {https://arxiv.org/abs/2207.12598}
}

@article{nguyen2023expresso,
  title={Expresso: A benchmark and analysis of discrete expressive speech resynthesis},
  author={Nguyen, Tu Anh and Hsu, Wei-Ning and d'Avirro, Antony and Shi, Bowen and Gat, Itai and Fazel-Zarani, Maryam and Remez, Tal and Copet, Jade and Synnaeve, Gabriel and Hassid, Michael and others},
  journal={arXiv preprint arXiv:2308.05725},
  year={2023}
}

\end{document}